\documentclass[10pt,twocolumn,letterpaper]{article}

\usepackage[pagenumbers]{cvpr} 

\usepackage{graphicx}
\usepackage{amsmath}
\usepackage{amssymb}
\usepackage{booktabs}
\usepackage{balance}

\usepackage[pagebackref,breaklinks,colorlinks]{hyperref}

\usepackage[capitalize]{cleveref}
\crefname{section}{Sec.}{Secs.}
\Crefname{section}{Section}{Sections}
\Crefname{table}{Table}{Tables}
\crefname{table}{Tab.}{Tabs.}

\usepackage{makecell,rotating}
\usepackage{multirow}
\usepackage{booktabs}
\usepackage{bbding}
\usepackage{pifont}
\usepackage{mathrsfs}
\usepackage{pythonhighlight}
\lstnewenvironment{PythonB}[1][]{\lstset{style=mypython, frame=none, breaklines=true, #1}}{}
\usepackage[ruled,boxed]{algorithm2e}

\newcommand{\repeatthanks}{\textsuperscript{\thefootnote}}

\def\cvprPaperID{*****} 
\def\confName{CVPR}
\def\confYear{2023}

\begin{document}

\title{TeachCLIP: Multi-Grained Teaching for Efficient Text-to-Video Retrieval}

\author{Kaibin Tian\textsuperscript{1}\thanks{Equal contribution.}, Ruixiang Zhao\textsuperscript{1}\repeatthanks, Hu Hu\textsuperscript{2}, Runquan Xie\textsuperscript{2}, Fengzong Lian\textsuperscript{2}, Zhanhui Kang\textsuperscript{2} and Xirong Li\textsuperscript{1}\thanks{Corresponding author: Xirong Li (xirong@ruc.edu.cn)}\\
\textsuperscript{1}{MoE Key Lab of DEKE, Renmin University of China} \\
\textsuperscript{2}{Tencent}
}
\maketitle

\begin{abstract}

For text-to-video retrieval (T2VR), which aims to retrieve unlabeled videos by ad-hoc textual queries, CLIP-based methods are dominating. Compared to CLIP4Clip which is efficient and compact, the state-of-the-art models tend to compute video-text similarity by fine-grained cross-modal feature interaction and matching, putting their scalability for large-scale T2VR into doubt. For efficient T2VR, we propose \emph{TeachCLIP} with multi-grained teaching to let a CLIP4Clip based student network learn from more advanced yet computationally heavy models such as X-CLIP, TS2-Net and X-Pool . To improve the student's learning capability, we add an Attentional frame-Feature Aggregation (AFA) block, which by design adds no extra storage / computation overhead at the retrieval stage. While attentive weights produced by AFA are commonly used for combining frame-level features, we propose a novel use of the weights to let them imitate frame-text relevance estimated by the teacher network. As such, AFA provides a fine-grained learning (teaching) channel for the student (teacher). Extensive experiments on multiple public datasets justify the viability of the proposed method. 
\end{abstract}


\section{Introduction}
\label{sec:introduction}

This paper aims for efficient text-to-video retrieval (T2VR), letting a common user efficiently retrieve specific videos from many unlabeled videos by textual queries. The T2VR task is challenging due to its cross-modal nature: videos and queries have to be encoded into a \emph{semantically}-aligned common feature space for video-text matching \cite{dualencoding}. 

\begin{figure}[t!]
  \centering
  \includegraphics[width=\columnwidth]{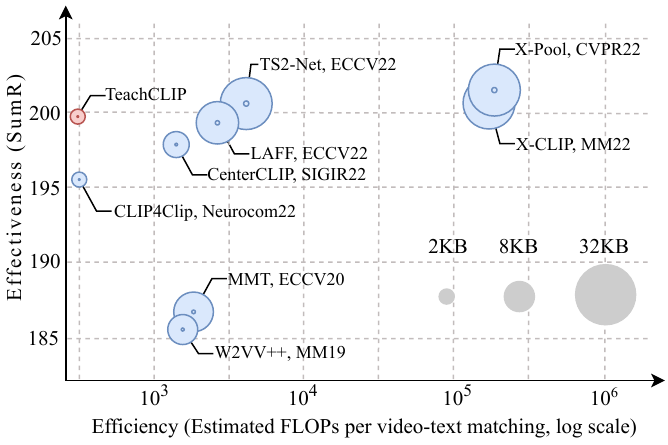}
  \caption{\textbf{Effectiveness, efficiency and video-feature storage footprint of present-day (CLIP based) text-to-video retrieval models}. Dataset: MSRVTT-1k \cite{msrvtt-1ka}. Backbone: CLIP(ViT-B/32) \cite{clip}. We propose TeachCLIP to strike a good balance between the three factors.}
  \label{fig:insight}
\end{figure}

Due to the great success of the Contrastive Language-Image Pre-Training (CLIP) model \cite{clip} in the image domain, we see encouraging efforts on re-purposing CLIP for video-text matching \cite{clip4clip,clip2video,xclip,ts2net,xpool}. As the first work in this line of research, CLIP4Clip \cite{clip4clip} encodes a given video by first using CLIP's visual encoder to extract image features per frame. The frame-level features, enhanced further by stacked Transformer blocks, are aggregated into a video-level feature by mean pooling. The video feature, with a typical dimension of 512, can be computed and stored in advance, making CLIP4Clip efficient for T2VR. Recent methods, \eg X-CLIP \cite{xclip}, TS2-Net \cite{ts2net} and X-Pool \cite{xpool}, improve over CLIP4Clip by considering more fine-grained frame-text similarities. Despite their better retrieval performance on popular benchmarks \cite{msrvtt,msvd}, these models introduce substantial overhead \wrt offline storage and online computation, see Fig. \ref{fig:insight}, putting their scalability for large-scale T2VR into question.

The research question arises as \emph{how to reduce the performance gap between CLIP4Clip and the latest CLIP-based methods}? 
A relatively straightforward approach is to apply knowledge distillation techniques \cite{hinton2015distilling}, which are developed for improving the performance of a student model by transferring dark knowledge from a relatively larger and stronger teacher model to the student. Indeed, there has been good effort on exploiting such techniques for T2VR. TeachText \cite{teachtext} first trains an ensemble of T2VR models with varied textual encoders. The averaged video-text similarity given by the ensemble is then used as soft labels to supervise a specific student network. As TeachText is designed specifically for video-level knowledge distillation, it remains unclear how to effectively pass fine-grained similarities, which are crucial for the teacher's good performance, on to the student?


On the other hand, the student network itself shall be capable of accepting both video-level and frame-level cross-modal knowledge from the teacher network. Given that CLIP4Clip lacks such capability, a specific module that can be supervised with both video-level and frame-level soft labels is needed. Meanwhile, the module shall not bring in extra storage / computation cost at the retrieval stage. Bearing these in mind, we propose an Attentional frame-Feature Aggregation (AFA) block, see Fig. \ref{fig:framework} for a conceptual diagram. Given a sequence of frame-level features as input, the AFA block produces frame-specific weights, which will be used to aggregate the frame features to a video-level feature.  Note that the block is not new by itself, as similar blocks have been used for fusing diverse video / text features \cite{laff}. What is novel here is that we not only use the attentive weights to weigh the frame features, but also let the weights imitate the frame-text relevance estimated by the teacher network. By doing so, AFA provides a fine-grained learning (teaching) channel for the student (teacher). In our setting, the student and the teacher are both CLIP-based, so the proposed method is coined \texttt{TeachCLIP}.


In sum, our main contributions are as follows:
\begin{itemize}
    \item We present a first study on reducing the performance gap between CLIP4Clip and the latest CLIP-based methods for efficient T2VR.
    \item We propose TeachCLIP with multi-grained teaching to let a CLIP4Clip based student network to learn from more advanced yet computationally heavy models such as X-CLIP, TS2-Net and X-Pool. In particular, we add the AFA block to CLIP4Clip, which by design introduces no extra storage / computation overhead at the retrieval stage.
    \item Extensive experiments on multiple datasets, \ie MSRVTT-1k \cite{msrvtt-1ka}, MSRVTT-3k \cite{msrvtt}, MSVD \cite{msvd}, VATEX \cite{vatex} and ActNetCap \cite{activitynet}, justify the viability of the proposed method. 
\end{itemize}

\section{Related Work}
\label{sec:related}

At a high level, 
our idea of multi-grained teaching can be 
viewed as knowledge distillation from a relatively heavy teacher network to an efficient student network. Hence, we briefly review progress in T2VR and knowledge distillation, accordingly interpreting our novelty in such a joint context.


\subsection{Text-to-Video Retrieval} \label{ssec:related-video}

\textbf{Effective T2VR}. The majority of the literature is on effectiveness, aiming for better cross-modal matching networks that compute video-text similarity more accurately. Depending on whether video / text feature extractors are trained together with the cross-modal matching module, we categorize existing works into two groups. That is, feature re-learning methods and (CLIP based) end-to-end methods.

Feature re-learning methods typically employ pretrained 2D-CNNs for  \cite{w2vv++,dualencoding}, 3D-CNNs \cite{je,ce} or their combinations \cite{mmt,laff} to obtain an initial feature representation of a given video. Similarly, a given text is encoded either by nontrainable bag-of-words \cite{w2vv++} or by pre-trained text encoders including Word2Vec \cite{w2vv}, BERT \cite{w2vvpp-bert}, GPT \cite{teachtext}, \etc.  Feature re-learning is then performed to project the video and text features into a common latent space, wherein the video-text relevance can be measured in terms of their distance in the common space. While there is still room for improvement, \eg by adding more features with novel feature fusion blocks, see MMT \cite{mmt}, TeachText \cite{teachtext} and LAFF \cite{laff}, the performance of feature re-learning methods is largely bounded by the initial features.  

The advent of CLIP \cite{clip} and its application in the video domain has significantly reshaped the research landscape of T2VR.  CLIP-based end-to-end methods have shown superior performance to their feature re-learning based predecessors on multiple public datasets \cite{clip4clip,ts2net,xpool,xclip}. As an initial attempt in this line of research, CLIP4Clip \cite{clip4clip} employs the visual encoder of CLIP to first extract a sequence of frame-level features. The frame features, updated by a stack of standard Transformer blocks, are averaged to produce a video-level feature. Such a video feature can be precomputed offline, while the storage footprint is linear \wrt the number of videos. Hence, a CLIP4Clip based T2VR system is efficient and compact. Follow-ups of CLIP4Clip, \eg X-CLIP \cite{xclip}, TS2-Net \cite{ts2net} and X-Pool \cite{xpool}, improve video-text matching by fine-grained cross-modal feature interaction and matching. Despite their better performance, local interaction means features used for cross-modal matching have to be computed online, while fine-grained matching results in substantial computation and storage overhead. This puts the scalability of the latest CLIP-based methods into question.

\textbf{Efficient T2VR}. In order to accelerate video feature extraction, CenterCLIP \cite{centerclip} utilizes a multi-segment token clustering algorithm to find the most representative tokens. Only these essential tokens will be forwarded through the entire Transformers, while those non-essential tokens will be dropped at a pre-specified Transformer block. For video-text relevance estimation, the model requires fine-grained matching at the segment level, making its storage and computation amplified by the number of segments per video. Hence, CenterCLIP remains relatively heavy when compared to CLIP4Clip. We aim for the same efficiency and compactness as CLIP4Clip at the retrieval stage.

\subsection{Knowledge Distillation} \label{ssec:related-knowledge}

Knowledge distillation is to transfer ``dark'' knowledge from a large, heavily weighted teacher model or ensemble of teacher models to a single, smaller student model, which can be practically deployed \cite{hinton2015distilling}. The form of the knowledge varies, which can be the output of the teacher \cite{peason}, its intermediate representations \cite{fitnets}, or mutual relations of data examples \cite{relation}.  In the context of video-to-video retrieval, DnS is proposed to distill knowledge, represented in the form of video-to-video relevance scores, from a big network to a smaller network \cite{dns}. As for T2VR, TeachText is developed \cite{teachtext}, where a student network is trained to mimic the averaged video-text similarity given by an ensemble of feature re-learning based models. Technically different from DnS and TeachText, the proposed TeachCLIP goes one step further by exploiting intermediate results of an up-to-date CLIP-based model as side information to supervise a CLIP4Clip based student model.





\section{Proposed TeachCLIP Method}

\begin{figure*}[t!]
  \centering  
  \includegraphics[width=2\columnwidth]{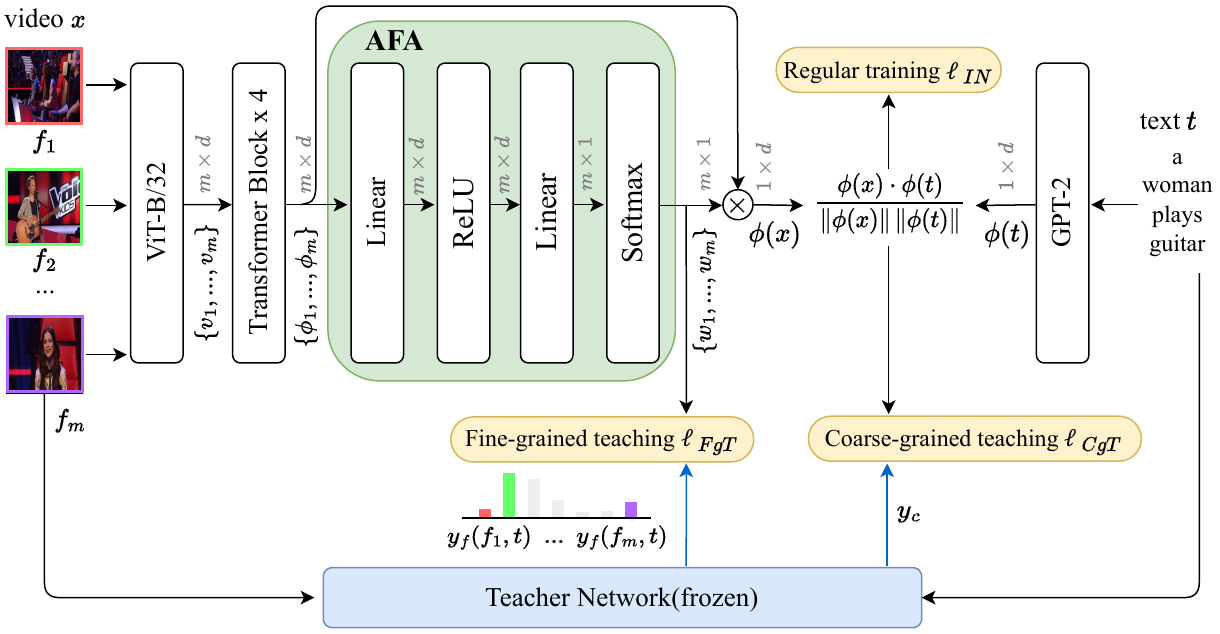}
  \caption{\textbf{Proposed \textit{TeachCLIP} method for training a video CLIP for efficient text-to-video retrieval}. Given a sequence of frame-level features as input, the AFA block produces frame-specific weights $\{w_1, \ldots, w_m\}$, used to aggregate the frame features to a video-level feature.  Meanwhile, by letting the weights imitate the frame-text relevance $\{y_f(f_1,t), \ldots, y_f(f_m,t)\}$ provided by the teacher network, AFA provides a fine-grained learning (teaching) channel for the student (teacher). The student network is end-to-end trained by jointly minimizing a coarse-grained teaching loss $\ell_{CgT}$, a fine-grained teaching loss $\ell_{FgT}$ and a regular training loss $\ell_{IN}$.}
  \label{fig:framework}
\end{figure*}

\subsection{Problem Setup} \label{ssec:setup}

The setup of this research is as follows.  We are provided with two sorts of CLIP-based T2VR models. The first sort,  using exclusively video-level features for video-text matching, is efficient in terms of storage footprint and retrieval speed. Meanwhile, the other sort, relying on fine-grained cross-modal matching, is more accurate than the former. Such an advantage, however, comes at the cost of substantially increased storage and computation overhead. Viewing the former as a student and the latter as a teacher, the proposed TeachCLIP method aims to improve the relatively weaker student with multi-grained dark knowledge from the teacher, see Fig. \ref{fig:framework} for a conceptual diagram. In what follows, we detail the student network in Sec. \ref{ssec:student}, which will be trained by the proposed multi-grained teaching algorithm in Sec. \ref{ssec:mgt}.




\subsection{The Student Network} \label{ssec:student}

Our student network is based on CLIP4Clip \cite{clip4clip}. Given a video $x$ represented by a sequence of $m$ frames $\{f_1, \ldots, f_m\}$, CLIP4Clip first feeds the frames in parallel into the visual encoder of CLIP, \ie a Vision Transformer (ViT), producing an array of frame-level features $\{v_1, \ldots, v_m\}$, sized $m \times d$.  These features, appended with position encoding, then go through four stacked Transformer blocks for temporal modeling, resulting in $m$ enhanced features $\{\phi_1, \phi_2, \ldots, \phi_m\}$. The video feature $\phi(x)$ is obtained by mean pooling over the enhanced features. As aforementioned, mean pooling, treating the individual frames equally, leads to the performance gap between CLIP4Clip and  recent methods (X-CLIP, TS2-Net, X-Pool \etc) that calculate fine-grained cross-modal similarities. 

We improve CLIP4Clip by replacing its mean pooling layer with an Attentional frame-Feature Aggregation (AFA) block. Given $\{\phi_1, \phi_2, \ldots, \phi_m\}$ as its input, AFA is designed to produce an $m$-dimensional nonnegative weight vector $\{w_1, \ldots, w_m\}$, where $w_i$ shall reflect the importance of frame $f_i$. More formally, the key data flow of the visual side of the student network is expressed as follows:
\begin{equation}
\left\{ \begin{array}{ll}
 \{f_1, \ldots, f_m\} & \leftarrow \mbox{video-to-frames}(x),\\
 \{v_1, \ldots, v_m\} & \leftarrow \mbox{ViT}(\{f_1, \ldots, f_m\}),\\
 \{\phi_1, \ldots, \phi_m\} & \leftarrow \mbox{Transformers}\times4(\{v_1, \ldots, v_m\}),\\
 \{w_1, \ldots, w_m\} & \leftarrow \mbox{AFA}(\{\phi_1, \ldots, \phi_m\}), \\
  \phi(x) & \leftarrow  \sum_{i=1}^m w_i \phi_i.
       \end{array} \right.
\end{equation}

As illustrated in \cref{fig:framework}, we implement the AFA block with a linear layer of $d\times d$, followed by ReLU, another linear layer of $d\times 1$, and finally a softmax layer. As such, the amount of extra parameters introduced by AFA is $O(d^2)$. Such  computational overhead is ignorable. More importantly, AFA provides a fine-grained learning path for the student network, as we will see shortly in \cref{sssec:fgt}.

\subsection{Multi-grained Teaching} \label{ssec:mgt}


The proposed multi-grained teaching (MgT) algorithm follows a standard SGD procedure. Per training iteration, a mini-batch $B$ of $b$ video-text pairs $\{(x_i, t_i) |i=1,\ldots,b\}$ is randomly sampled from a given training dataset. Such randomness allows us to consider each video  irrelevant \wrt other texts and vice versa within the batch. To simplify our notation, we re-use $B$ as a $b\times b$ video-text similarity matrix derived from the batch by the student network. More specifically, we have $B_{i,j}$ as the cosine similarity between $\phi(x_i)$ and $\phi(t_j)$, $i,j=1,\ldots,b$. Accordingly, the $i$-th row $B_{i,\cdot}$ stores similarity scores of video $x_i$ \wrt all $b$ texts in the batch, while the $j$-th column $B_{\cdot,j}$ stores similarity scores of all $b$ videos \wrt text $t_j$. 

In order to perform MgT, we require two outputs from the teacher network concerning video $x_i$ and text $t_j$. That is, a coarse-grained relevance score of $x_i$ \wrt $t_j$, denoted by $y_c(x_i,t_j)$, and a fine-grained relevance score per frame, denoted by $y_f(f_{i,k},t_j)$, $k=1,\ldots,m$. Note that $y_c$ and $y_f$ can be obtained with ease from the SOTA models such as X-CLIP and TS2-Net that compute both frame-level and video-level relevance scores. Moreover, we assume that $\{y_f(f_{i,k},t_j)\}$ have been softmax-adjusted.

\subsubsection{Coarse-grained Teaching}

Coarse-grained teaching is to supervise the student network with video-level soft labels predicted by the teacher network. To that end, TeachText \cite{teachtext} uses the element-wise Huber loss, enforcing $B_{i,j}$ to be close to $y_c(x_i, t_j)$. However, recent research on knowledge distillation \cite{peason} suggests such type of loss is suboptimal, as enforcing a student model to replicate the output of a much stronger teacher model would unnecessarily increase the difficulty of knowledge distillation and consequently impede transferring truly useful knowledge from the teacher to the student. Instead, \cite{peason} suggests to minimize the Pearson's distance $d_p$ (or equivalently maximize Pearson correlation coefficient), which is known to be invariant under separate changes in scale and location in two given variables.  We consider such invariance also desirable in the current task, as minimizing $d_p$ between the output of the student and the teacher networks is adequate for the student to rank-wisely imitate the teacher. Following this thought, we opt to use $d_p$ as the coarse-grained teaching loss $\ell_{CgT}$. In particular, the loss for video $x_i$ is computed as $d_p(\sigma(B_{i,\cdot}), \sigma(y_c(v_i, \cdot)))$, where $\sigma$ is softmax. In a similar manner, the loss for text $t_j$ is calculated as $d_p(\sigma(B_{\cdot, j}), \sigma(y_c(\cdot, t_j))$. Accordingly, $\ell_{ct}$ is defined as the following batch-level symmetric loss:
%
%
\begin{equation} \label{eq:loss-cgt}
\begin{array}{ll}
 \ell_{CgT}  := & \frac{1}{b} \sum_{i=1}^b d_p(\sigma(B_{i,\cdot}), \sigma(y_c(v_i, \cdot)) + \\ 
 & \frac{1}{b} \sum_{j=1}^b d_p(\sigma(B_{\cdot, j}), \sigma(y_c(\cdot, t_j)) ). 
\end{array}
\end{equation}

\subsubsection{Fine-grained Teaching} \label{sssec:fgt}

Recall that at the visual side of the student network, we introduce a lightweight AFA  block to produce $m$ attention weights $\mathbf{w}=\{w_1, \ldots, w_m\}$ for convex combination of the $m$ temporally enhanced frame-level features $\{\phi_1, \ldots, \phi_m\}$. Instead of letting the network learn the weights all by itself, we propose to guide the weights-related learning process with fine-grained relevance information from the stronger teacher network. Intuitively, a frame that is more relevant \wrt the given text shall be weighed more. We therefore compute a fine-grained teaching loss $\ell_{FgT}$ per relevant video-text pair $(v_i, t_i)$ as the cross entropy (CE) loss between the weights and the teacher provided frame-text similarities. A batch-level loss is thus obtained by averaging over all the $b$ relevant pairs in the given batch, namely
\begin{equation} \label{eq:loss-fgt}
\ell_{FgT} := -\frac{1}{b}\sum_{i=1}^b \sum_{k=1}^m y_f(f_{i,k}, t_i) \log w_{i,k}.
\end{equation}

Notice that we do not consider fine-grained teaching on the textual side, \eg exploiting video-word similarities to attentively aggregate word-level features. Our main concern is that in contrast to a keyframe that represents the video content to a large extent, a keyword token alone is largely insufficient to represent the corresponding sentence. Indeed, our preliminary experiment showed that adding an AFA on the textual side brings no improvement. We therefore did not go further in that direction. Following CLIP4Clip, we adopt CLIP's textual encoder, \ie a GPT-2, without changing its network structure. 


In addition to $\ell_{CgT}$ and $\ell_{FgT}$, we calculate the symmetric InfoNCE loss over the similarity matrix $B$. Denoted as $\ell_{IN}$, this loss is commonly used for training cross-modal matching networks \cite{xclip,ts2net,clip4clip,centerclip}. The student network is trained to minimize the sum of the three losses, \ie $\ell_{CgT} + \ell_{FgT} + \ell_{IN}$, with the first two terms responsible for MgT and the last term for regular training. TeachCLIP is easy to implement, see Algorithm \ref{alg:teachclip}.

\begin{algorithm}
\caption{TeachCLIP in a PyTorch style}
\label{alg:teachclip}

\begin{PythonB}
Input: Training data loader D={(v, t)}
       Trained teacher network
Ouptput: Trained student network

optimizer = torch.optim.Adam(student.parameters)
for e=1,2,..., MAX_EPOCHES:
    for mini-batch {(v, t)} in D:
        optimizer.zero_grad()
        y_c, y_f = teacher({(v, t)}) 
        B, w = student({(v, t)})
        l_CgT = pearson_distance_loss(y_c, B)
        l_FgT = cross_entropy_loss(y_f, w)
        l_IN = symmetric_InfoNCE_loss(B)
        loss = l_CgT + l_FgT + l_IN
        loss.backward()
        optimizer.step()
\end{PythonB}
\end{algorithm}




    


\section{Experiments}
\label{sec:experiment}


\subsection{Experimental Setup} 

\textbf{Datasets}. We adopt the following public datasets that have been frequently used for T2VR evaluation: 
MSRVTT\cite{msrvtt}, MSVD\cite{msvd}, VATEX\cite{vatex} and ActivityNet-Caption (ActNetCap) \cite{activitynet}. While the original data split of MSRVTT has nearly 3k test videos and 60k sentences, Yu \etal suggest another split of 9k videos for training and 1k video-text pairs for testing \cite{msrvtt-1ka}. Probably due to the relatively smaller testset size that makes the evaluation more computation-resource friendly, the 1k edition (MSRVTT-1k) appears to be more popular than its 3k counterpart (MSRVTT-3k). We follow this practice, using MSRVTT-1k as the primary dataset for our ablation study. 

For MSVD, we use its official data slit. For VATEX, we use the split by Chen \etal \cite{hrg}. As for ActNetCap, we adopt the split by Gabeur \etal \cite{mmt}, testing on `val1' as \cite{clip4clip,xclip,ts2net}. Also notice that in contrast the other datasets that use a sentence as a query, ActNetCap, with descriptions per video merged into a paragraph, essentially performs paragraph-to-video retrieval. See \cref{tab:datasets} for an overview.

\begin{table}[ht!]
\centering
\caption{\textbf{Experimental data}. MSRVTT-1k is used as the primary dataset for ablation study, while all datasets are used to compare TeachCLIP with existing methods.}
\label{tab:datasets}
\resizebox{\linewidth}{!}{
\begin{tabular}{@{}lrrrrrr@{}}
\toprule
\multirow{2}{*}{\textbf{Dataset}} & \multicolumn{2}{c}{\textbf{Training set}} & \multicolumn{2}{c}{\textbf{Validation set}} & \multicolumn{2}{c}{\textbf{Test set}} \\ \cmidrule(r){2-3} \cmidrule(r){4-5} \cmidrule(r){6-7}
 & \#videos & \#texts & \#videos & \#texts & \#videos & \#texts \\ \hline
MSRVTT-1k \cite{msrvtt-1ka} & 9,000 & 180,000 & n.a. & n.a. & 1,000 & 1,000 \\
MSRVTT-3k \cite{msrvtt} & 6,513 & 130,260 & 497 & 9,940 & 2,990 & 59,800 \\
MSVD \cite{msvd} & 1,200 & 48,774 & 100 & 4,290 & 670 & 27,763 \\
VATEX \cite{vatex} & 25,991 & 259,910 & 1,500 & 15,000 & 1,500 & 15,000 \\
ActNetCap \cite{activitynet} & 10,009 & 10,009 & n.a. & n.a. & 4,917 & 4,917 \\
\bottomrule
\end{tabular}
}
\end{table}







\textbf{Evaluation criteria}.
We report standard rank-based metrics, \ie Recall at top $k$ ($k$=1, 5, 10) and SumR (R1+R5+R10) as a combined metric to assess the overall performance of a specific method.

\textbf{Implementation details}. Subject to our computation capacity ($8\times$ NVIDIA GeForce RTX 3090 24GB GPUs), the default setting is as follows, unless otherwise specified. We use ViT-B/32 as the visual encoder and GPT-2 as the textual encoder, initialized using an OpenAI-released CLIP\footnote{\url{https://github.com/openai/CLIP}}.  In order to reduce the risk of catastrophic forgetting, we empirically set the initial learning rate in a module-specific manner: 1e-7 for ViT-B/32 and GPT-2, and 1e-4 for the remaining modules in a specific network. The network is trained in 10 epochs at maximum by an Adam optimizer \cite{adam}, with the learning rate decayed by a cosine schedule strategy \cite{sgdr}. The input frame size is $224 \times 224$. The maximum length of frame tokens and word tokens are set to 12 and 32, respectively. The mini-batch size is 240, except for ActNetCap, which has much longer videos and texts than the other datasets and thus a smaller batch size of 96 is used. 

For MSRVTT-3k, MSVD and VATEX each of which has a separate validation set,  models maximizing R1 on the validation set are chosen. As for MSRVTT-1k and ActNetCap which has no such validation set, we follow \cite{clip4clip,xclip,ts2net}, reporting peaking performance on the test set.




\subsection{Evaluating TeachCLIP} \label{ssec:ablation}


Recall that the essence of TeachCLIP is to let a stronger teacher network to teach a computationally efficient student network for better retrieval performance. Hence, TeachCLIP needs to be evaluated along multiple dimensions including the choice of the teacher, the choice of the student, and how the teaching process is executed.

\textbf{Choice of the teacher network}.  We have X-CLIP\footnote{\url{https://github.com/xuguohai/X-CLIP}} \cite{xclip}, TS2-Net\footnote{\url{https://github.com/yuqi657/ts2_net}} \cite{ts2net} and X-Pool\footnote{\url{https://github.com/layer6ai-labs/xpool}} \cite{xpool} in our shortlist, as they are open-source, provide both video-text and frame-text similarity scores, and report leading performance on multiple datasets. Our experiments show that the performance of TS2-Net is somewhat unstable: better than CLIP4Clip on MSRVTT-1k (SumR 200.5 versus 195.5) and MSRVTT-3k (153.6 versus 150.1), yet worse on MSVD (204.9 versus 206.6) and ActNetCap (190.4 versus 194.1). As for X-Pool, we find that its official code requires computing a big cross-modal similarity matrix between all test videos and texts on a single GPU. Consequently, X-Pool is computationally feasible on MSRVTT-1k only. As such, we choose X-CLIP as the default teacher, unless otherwise stated.

As shown in Table \ref{table:ablation},  the student network \emph{without} teaching (Setup\#2) scores SumR of 196.3. By the proposed multi-grained teaching (MgT), the student has improved consistently, obtaining better SumR, see Setup\#7 (199.8, taught by X-CLIP), Setup \#11 (199.4, taught by TS2-Net), and Setup \#12 (200.1, taught by X-Pool). Comparing the performance of a specific teacher and its student, we see a clear pattern that their performance is highly correlated. Better teacher leads to better student. 

We also try a multi-teacher teaching strategy (Setup\#13), where $\ell_{CgT}$ and $\ell_{FgT}$ are computed per teacher and minimized jointly. Compared to its single-teacher counterparts, \ie Setup\#7 and Setup\#11, better performance is obtained, yet with no extra overhead in the retrieval stage.

In addition, we experiment with a stronger backbone, substituting ViT-B/16 for ViT-B/32 as the visual encoder. While both teacher, \ie X-CLIP (ViT-B/16), and student (Setup\#1) are better than their ViT-B/32 counterparts, MgT is again effective, lifting SumR from 201.2 to 207.0. 



\begin{table}[tb!]
\centering
\caption{\textbf{Evaluating TeachCLIP in varied setups}. For the ease of reference, performance of specific teacher models is also included. Dataset: MSRVTT-1k.}
\label{table:ablation}
\renewcommand{\arraystretch}{1.1} 
\resizebox{\linewidth}{!}{
\begin{tabular}{@{}lrrrr@{}}
\hline
\textbf{Model} & \textbf{R1} & \textbf{R5} & \textbf{R10} & \textbf{SumR} \\ \hline

\multicolumn{5}{@{}l}{\textit{Teacher:}} \\
X-CLIP \cite{xclip} & 45.3 & 73.7 & 81.8 & 200.8 \\
TS2-Net \cite{ts2net} & 46.7 & 72.6 & 81.2 & 200.5 \\
X-Pool \cite{xpool} & 46.0 & 72.8 & 82.7 & 201.5 \\
X-CLIP (ViT-B/16) & 49.1 & 76.1 & 84.4 & 209.6 \\ [2pt]
\multicolumn{5}{@{}l}{\textit{Student:}} \\
0: CLIP4Clip \cite{clip4clip} & 42.8 & 71.6 & 81.1 & 195.5 \\
1: CLIP4Clip(ViT-B/16) & 45.4 & 73.5 & 82.3 & 201.2 \\
2: \#0 + AFA & 44.0 & 71.2 & 81.1 & 196.3 \\ [2pt]
\multicolumn{5}{@{}l}{\textit{Student, taught by X-CLIP:}} \\
3: \#0 + CgT & 44.2 & 71.7 & 80.6 & 196.5 \\
4: \#2 + CgT & 44.4 & 71.0 & 82.6 & 198.0 \\
5: \#2 + CgT (Huber as $\ell_{CgT}$) & 44.7 & 71.1 & 82.0 & 197.8 \\
6: \#2 + FgT & 44.1 & 71.2 & 82.0 & 197.3 \\
7: \#2 + MgT & 45.2 & 72.3 & 82.3 & 199.8 \\
8: \#2 + MgT(CE as $\ell_{FgT}$) & 45.2 & 72.3 & 82.3 & 199.8 \\
9: \#2 + MgT(Pearson as $\ell_{FgT}$) & 45.4 & 73.0 & 81.3 & 199.7 \\
10: \#2 + MgT(Huber as $\ell_{FgT}$) & 45.6 & 71.9 & 81.8 & 199.3 \\ [2pt]

\multicolumn{5}{@{}l}{\textit{Student, taught by others:}} \\
11: \#7, by TS2-Net & 45.6 & 72.1 & 81.7 & 199.4 \\
12: \#7, by X-Pool & 44.0 & 73.5 & 82.6 & 200.1 \\
13: \#7, by X-CLIP \& TS2-Net  & 44.8 & 73.5 & 82.9 & 201.2 \\ 
14: \#7, by X-CLIP(ViT-B/16) & 48.1 & 75.4 & 83.5 & 207.0 \\
\hline
\end{tabular}
}
\end{table}



\textbf{Choice of the student network}. The better performance of Setup\#2 against Setup\#0 (196.3 versus 195.5) indicates that the AFA module is helpful, even under regular training. Further, we compare CLIP4Clip and CLIP4Clip+AFA in the coarse-grained teaching (CgT) mode, see Setup\#3 and Setup\#4. The latter has higher SumR of 198.0. Our choice of using CLIP4Clip+AFA as the student network is verified.


\textbf{Which loss for CgT}? Comparing Setup\#4 (the Pearson distance loss $d_p$) and Setup\#5 (the Huber loss as previously used by TeachText), the former is marginally better (198.0 versus 197.8). In fact, our experiments on the other datasets show that $d_p$ consistently outperforms the Huber loss. 


\textbf{Is MgT really necessary}?  Setup\#4, Setup\#6 and Setup\#7 correspond to  the student network trained by CgT, by FgT and by MgT, respectively. MgT has SumR of 199.8, followed by CgT (198.0) and FgT (197.3). With MgT, the performance gap between the student (CLIP4Clip+AFA) and the teacher (X-CLIP), measured by the absolute difference of SumR, is reduced from 4.5 to 1.0. The necessity of MgT is verified.

\textbf{Which loss for FgT}? By default, we use CE as $\ell_{FgT}$. We compare CE with two alternatives, namely $d_p$ and Huber, see Setup\#8 to \#10. CE is the best loss for FgT.

\begin{figure*}[ht!]
  \centering
  \includegraphics[width=2\columnwidth]{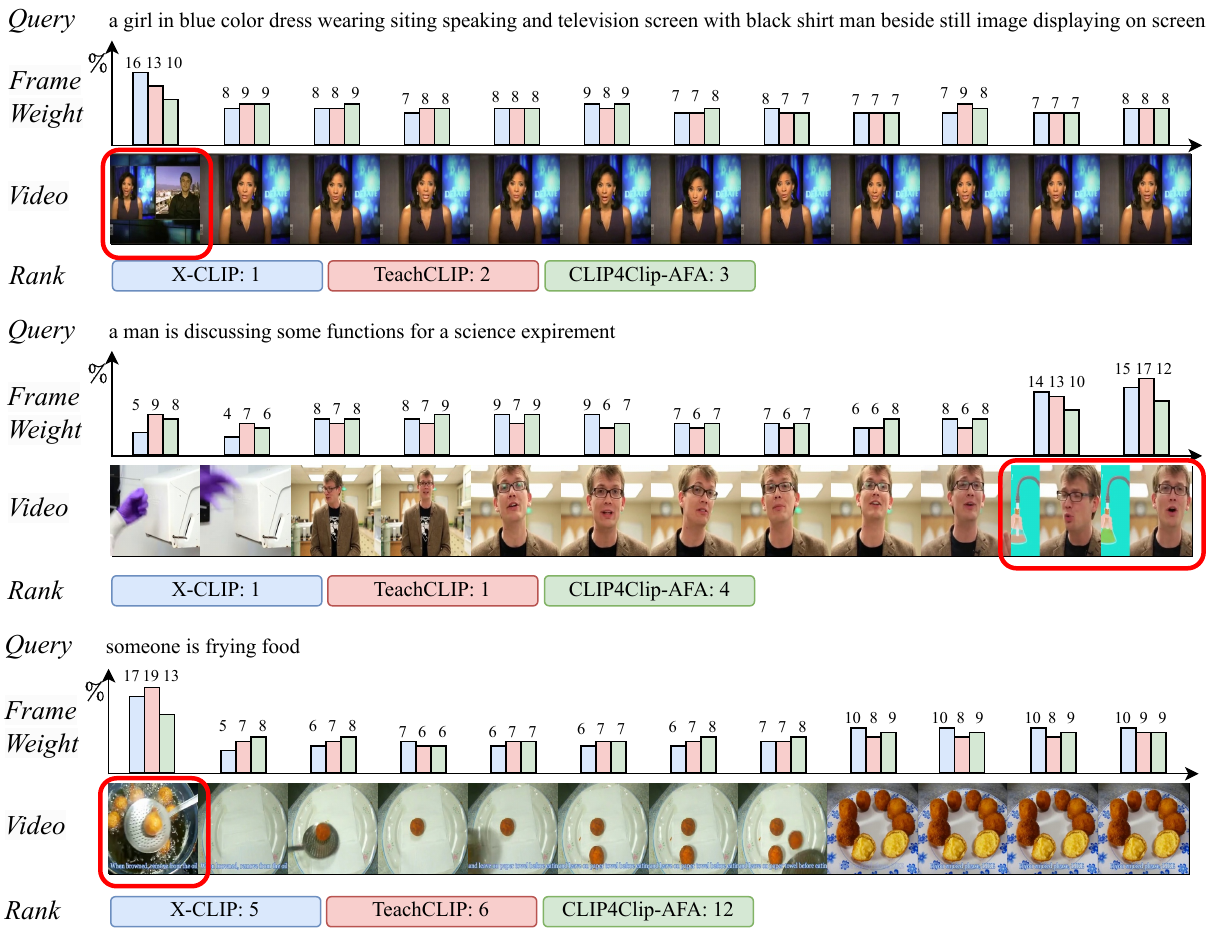}
  \caption{\textbf{Visualization of video frames,  frame relevance / weights by X-CLIP (teacher),  CLIP4Clip-AFA (student w/o MgT) and TeachCLIP ( student with MgT), and video ranks} (lower rank better). While both students have their frame weights pre-computed in a query-independent manner, the weights by TeachCLIP are more close to the query-dependent weights by X-CLIP, especially on salient frames (manually marked out by red rectangles for the ease of reference). 
  }
  \label{fig:vis}
\end{figure*}

\textbf{Qualitative analysis}. Some qualitative results are given in Fig. \ref{fig:vis}. Consider the result at the top for instance. The first frame is the most salient, as it shows key objects, \ie a girl in blue color dress and black shirt man, specified in the query. By computing frame-text relevance on the fly, X-CLIP successfully identifies this frame. For this frame, TeachCLIP also gives a larger weight, albeit precomputed. Similar results can be observed in the other two examples. These qualitative results further confirm the viability of TeachCLIP.



\subsection{Comparison with Existing Methods} \label{ssec:performace}

\textbf{Baselines}. We compare with both feature re-learning based methods and CLIP-based end-to-end methods. For the purpose of reproducible research, we include the following peer-reviewed, open-sourced methods: 
\begin{itemize}
    \item \emph{Feature re-learning}: W2VV++\footnote{\url{https://github.com/li-xirong/w2vvpp}}\cite{w2vv++}, DualE\footnote{\url{https://github.com/danieljf24/hybrid_space}}\cite{dualencoding}, CE\footnote{\url{https://github.com/albanie/collaborative-experts}}\cite{ce}, SEA\footnote{\url{https://github.com/li-xirong/sea}}\cite{sea}, MMT\footnote{\url{https://github.com/gabeur/mmt}}\cite{mmt}, TeachText\footnote{\url{https://github.com/albanie/collaborative-experts}}\cite{teachtext}, and LAFF\footnote{\url{https://github.com/ruc-aimc-lab/laff}}\cite{laff}. 
    \item \emph{CLIP-based end-to-end}: Besides CLIP4Clip \cite{clip4clip}, X-CLIP \cite{xclip}, TS2-Net \cite{ts2net} and X-Pool \cite{xpool} that have been evaluated in our ablation study, we include CenterCLIP\footnote{\url{https://github.com/mzhaoshuai/CenterCLIP}}\cite{centerclip}.
\end{itemize}


\begin{table*}[ht!]
\centering
\caption{\textbf{T2VR Performance of different methods on multiple datasets}. Note that we replicate existing methods with their author-provided source code where applicable, so the numbers might differ (slightly) from their original papers.
}
\label{table:alldatasets}
\renewcommand{\arraystretch}{1.1} 
\resizebox{\linewidth}{!}{
\begin{tabular}{@{}lrrrrrrrrrrrrrrrrrrrrrr@{}}
\toprule
\multirow{2}{*}{\textbf{Model}} & \multirow{2}{*}{\makecell[c]{FLOPs\\(k)$\downarrow$}} & \multirow{2}{*}{\makecell[c]{Store\\(KB)$\downarrow$}} & \multicolumn{3}{c}{\textbf{MSRVTT-1k}} &\multicolumn{3}{c}{\textbf{MSRVTT-3k}} & \multicolumn{3}{c}{\textbf{MSVD}} & \multicolumn{3}{c}{\textbf{VATEX}} & \multicolumn{3}{c}{\textbf{ActNetCap}} \\ 
\cmidrule(r){4-6} \cmidrule(r){7-9} \cmidrule(r){10-12} \cmidrule(r){13-15} \cmidrule(r){16-18}
 & & & R1 & R5 & SumR & R1 & R5 & SumR & R1 & R5 & SumR & R1 & R5 & SumR & R1 & R5 & SumR \\ \hline
\multicolumn{17}{@{}l}{\textit{Feature re-learning w/o CLIP feature}:}\\
W2VV++, MM19 \cite{w2vv++} & 2.0 & 8 & 18.9 & 45.3 & 121.7 & 11.1 & 29.6 & 81.2 & 22.4 & 51.6 & 138.8 & - & - & - & - & -  & - \\
DualE, PAMI21 \cite{dualencoding} & 2.0 & 8 & 21.1 & 48.7 & 130.0 & 11.6 & 30.3  & 83.2 & - & - & - & 36.8 & 73.6  & 194.1 & -  & - & - \\
CE, BMVC19 \cite{ce} & 6.1 & 27 & 20.9 & 48.8  & 132.1 & 10.0 & 29.0  & 80.2 & 19.8 & 49.0 & 132.6 & - & -  & - & 17.7 & 46.6 &  - \\
SEA, TMM21 \cite{sea} &  10.2 & 40 & 23.8 & 50.3  & 137.9 & 13.1 & 33.4  & 91.5 & 24.6 & 55.0  & 147.5 & - & -  & - & - & -  & - \\
MMT, ECCV20 \cite{mmt} & 3.6 & 14 & 24.6 & 54.0  & 145.7 & - & -  & - & - & - & - & - & -  & - & 22.7 & 54.2  & - \\
TeachText, ICCV21 \cite{teachtext} & 0.8 & 3 & 29.6 & 61.6  & 165.4 & 15.0 & 38.5  & 105.2 & 25.4 & 56.9  & 153.6 & 53.2 & 87.4  & 233.9  & 23.5 & 57.2  & - \\ [3pt]
\multicolumn{17}{@{}l}{\textit{Feature re-learning with CLIP feature}:}\\
SEA  & 10.2 & 40 & 37.2 & 67.1  & 182.6 & 19.9 & 44.3  & 120.7 & 34.5 & 68.8  & 183.8 & 52.4 & 90.2  & 238.5 & - & -  & - \\
W2VV++  & 2.0 & 8 & 39.4 & 68.1  & 185.6 & 23.0 & 49.0 & 132.7 & 37.8 & 71.0 &  190.4 & 55.8 & 91.2 & 243.0 & - & -  & - \\
MMT  & 3.6 & 14 & 39.5 & 68.3  & 186.1 & 24.9 & 50.5 &  137.4 & 40.6 & 72.0 & 194.3 & 54.4 & 89.2  & 238.6 & - & - & - \\
LAFF, ECCV22 \cite{laff} & 4.1 & 16 & 45.8 & 71.5  & 199.3 & 29.1 & 54.9  & 149.8 & 45.4 & 70.6 & 200.6 & 59.1 & 91.7 & 247.1 & - & - & -  \\  [3pt]
\multicolumn{17}{@{}l}{\textit{CLIP-based end-to-end}:}\\
CenterCLIP, SIGIR22\cite{centerclip} & 1.5$\sim$7.7 & 6$\sim$30 &  44.2 & 71.6 & 197.9 & - & - & -  & 47.3 & 76.8  & 209.7 & - & - & - & 43.9 & \textbf{74.6}  & \textbf{204.3} \\
CLIP4Clip, Neurocom22 \cite{clip4clip} & \textbf{0.5} & \textbf{2} & 42.8 & 71.6  & 195.5 & 29.4 & 54.9 & 150.1 & 45.6 & 76.1  & 206.6 & 61.6 & 91.1  & 248.5 & 39.7 & 71.0  & 194.1 \\
TS2-Net, ECCV22 \cite{ts2net} & 6.1$\sim$32.8 & 24$\sim$128 & \textbf{46.7} & 72.6  & 200.5 & 29.9 & 56.4  & 153.6 & 44.6 & 75.8  & 204.9 & 61.1 & 91.5  & 248.6 & 37.3 & 69.9  & 190.4 \\
X-CLIP, MM22 \cite{xclip} & 220.9$\sim$2175.9 & 26$\sim$130 & 45.3 & \textbf{73.7} & 200.8 & \textbf{31.2} & \textbf{57.4} &  \textbf{156.7} & 47.2 & 77.0  & 210.1 & 62.2 & 90.9  & 248.5 & \textbf{44.4} & \textbf{74.6}  & 204.1 \\
X-Pool, CVPR22 \cite{xpool} & 275.0 & 24 & 46.0 & 72.8 &  \textbf{201.5}  & - & - & - & - & - & - & - & - & - & - & - & - \\
[3pt]
TeachCLIP(X-CLIP) & \textbf{0.5} & \textbf{2} & 45.2 & 72.3 & 199.8 & 30.4 & 56.3 & 153.4 & \textbf{47.4} & \textbf{77.3} &  \textbf{210.2} & \textbf{63.6} & \textbf{91.9} &  \textbf{251.6} & 42.2 & 72.7 & 200.1  \\ 
\bottomrule
\end{tabular}
}
\end{table*}

\textbf{Effectiveness comparison}. The multi-dataset performance of the varied methods is given in Table \ref{table:alldatasets}. Noticeable performance gap exists between the feature re-learning based methods (the top part of Table \ref{table:alldatasets}) and the CLIP-based end-to-end methods (the bottom part of the same table). The former can be much improved by adding pre-extracted CLIP features (the middle part, cited from \cite{laff}). Consider MMT, for instance. The inclusion of the CLIP feature brings in a clear improvement (SumR 145.7 $\xrightarrow{}$ 186.1 on MSRVTT-1k). Nonetheless, they remain inferior to the end-to-end methods.

Among the end-to-end methods being compared, there is no clear winner that tops the performance on all the five datasets. X-Pool is the best on MSRVTT-1k, CenterCLIP is the best on ActNetCap, whilst X-CLIP leads on MSRVTT-3k and MSVD. Still, our evaluation confirms that X-CLIP has the best overall performance. 

The performance gap between CLIP4Clip and X-CLIP, measured in terms of the absolute difference of their SumR scores per dataset, is as follows: MSRVTT-1k 5.3, MSRVTT-3k 6.6, MSVD 3.5, VATEX 0.0, and ActNetCap 10.0. The overall gap averaged over the datasets is 5.1. 
TeachCLIP, as efficient as CLIP4Clip, reduces the above gap as follows:  MSRVTT-1k 5.3 $\rightarrow$ 1.0, MSRVTT-3k 6.6 $\rightarrow$ 3.3, and ActNetCap 10.0 $\rightarrow$ 4.0. Even more, TeachCLIP (marginally) surpasses X-CLIP on MSVD (210.2 versus 210.1) and VATEX (251.6 versus 248.5). While one would normally not  expect the student to beat the teacher, our interpretation of this counter-intuitive result is as follows. When the teacher and the student have  distinct  network structures yet with close performance, as in the cases of MSVD and VATEX, the teacher may provide complementary information that the student cannot learn by itself. As such, MgT has an effect of ensemble learning to train a better model. In sum, the overall gap is reduced from 5.1 to 1.0. The result justifies the effectiveness of the proposed TeachCLIP method.

\textbf{Storage comparison}. Per model, we calculate for a given video all features that are needed at the retrieval stage and can be precomputed. Consider X-CLIP and X-Pool for instance. Frame features are needed for frame-text matching, so they have to be stored. By contrast, CLIP4Clip and TeachCLIP use no frame features for retrieval. Given 4 bytes per floating point, CLIP4Clip and TeachCLIP have the smallest storage footprint of 2KB per video, see the third column of Table \ref{table:alldatasets}.

\textbf{Efficiency comparison}. For model, we measure the number of FLOPs\footnote{\url{https://github.com/Lyken17/pytorch-OpCounter}} required per video-text matching. As shown in the second column of Table \ref{table:alldatasets}, CLIP4Clip and TeachText are the most efficient.





\section{Conclusions}
\label{sec:conclusion}


We propose TeachCLIP with multi-grained teaching (MgT) for efficient text-to-video retrieval (T2VR). Extensive experiments on multiple public datasets allow us to conclude as follows.  The performance of the student and the teacher is positively correlated. Better teacher leads to better student. While coarse-grained teaching and fine-grained teaching are helpful even used separately, their joint use, namely MgT, is the best. The overall performance gap between CLIP4Clip (w/o teaching) and X-CLIP is 5.1 in terms of SumR averaged over all test sets. With MgT, the gap is reduced to 1.0. As the student network is designed to have the same efficiency and compactness as CLIP4Clip at the retrieval stage, the much reduced performance gap justifies the viability of the proposed method.

\textbf{Acknowledgments}. 
This work was supported by NSFC (62172420), Tencent Marketing Solution Rhino-Bird Focused Research Program, and Public Computing Cloud, Renmin University of China.

\balance
{\small
\bibliographystyle{ieee_fullname}
\bibliography{egbib}

\begin{thebibliography}{10}\itemsep=-1pt

\bibitem{msvd}
David Chen and William~B Dolan.
\newblock Collecting highly parallel data for paraphrase evaluation.
\newblock In {\em ACL}, pages 190--200, 2011.

\bibitem{hrg}
Shizhe Chen, Yida Zhao, Qin Jin, and Qi Wu.
\newblock Fine-grained video-text retrieval with hierarchical graph reasoning.
\newblock In {\em CVPR}, pages 10638--10647, 2020.

\bibitem{teachtext}
Ioana Croitoru, Simion-Vlad Bogolin, Marius Leordeanu, Hailin Jin, Andrew
  Zisserman, Samuel Albanie, and Yang Liu.
\newblock {TeachText}: Crossmodal generalized distillation for text-video
  retrieval.
\newblock In {\em ICCV}, pages 11583--11593, 2021.

\bibitem{w2vv}
Jianfeng Dong, Xirong Li, and Cees G.~M. Snoek.
\newblock Predicting visual features from text for image and video caption
  retrieval.
\newblock {\em TMM}, 20(12):3377--3388, 2018.

\bibitem{dualencoding}
Jianfeng Dong, Xirong Li, Chaoxi Xu, Xun Yang, Gang Yang, Xun Wang, and Meng
  Wang.
\newblock Dual encoding for video retrieval by text.
\newblock {\em TPAMI}, 44(8):4065--4080, 2021.

\bibitem{clip2video}
Han Fang, Pengfei Xiong, Luhui Xu, and Yu Chen.
\newblock {CLIP2Video}: Mastering video-text retrieval via image {CLIP}.
\newblock {\em Arxiv}, 2021.

\bibitem{mmt}
Valentin Gabeur, Chen Sun, Karteek Alahari, and Cordelia Schmid.
\newblock Multi-modal transformer for video retrieval.
\newblock In {\em ECCV}, pages 214--229, 2020.

\bibitem{xpool}
Satya~Krishna Gorti, No{\"e}l Vouitsis, Junwei Ma, Keyvan Golestan, Maksims
  Volkovs, Animesh Garg, and Guangwei Yu.
\newblock X-{P}ool: Cross-modal language-video attention for text-video
  retrieval.
\newblock In {\em CVPR}, pages 5006--5015, 2022.

\bibitem{activitynet}
Fabian~Caba Heilbron, Victor Escorcia, Bernard Ghanem, and Juan~Carlos Niebles.
\newblock {ActivityNet}: A large-scale video benchmark for human activity
  understanding.
\newblock In {\em CVPR}, pages 961--970, 2015.

\bibitem{hinton2015distilling}
Geoffrey Hinton, Oriol Vinyals, and Jeff Dean.
\newblock Distilling the knowledge in a neural network.
\newblock {\em arXiv}, 2015.

\bibitem{laff}
Fan Hu, Aozhu Chen, Ziyue Wang, Fangming Zhou, Jianfeng Dong, and Xirong Li.
\newblock Lightweight attentional feature fusion: A new baseline for
  text-to-video retrieval.
\newblock In {\em ECCV}, pages 444--461, 2022.

\bibitem{peason}
Tao Huang, Shan You, Fei Wang, Chen Qian, and Chang Xu.
\newblock Knowledge distillation from a stronger teacher.
\newblock In Alice~H. Oh, Alekh Agarwal, Danielle Belgrave, and Kyunghyun Cho,
  editors, {\em NeurIPS}, 2022.

\bibitem{adam}
Diederik~P Kingma and Jimmy Ba.
\newblock Adam: A method for stochastic optimization.
\newblock {\em arXiv}, 2014.

\bibitem{dns}
Giorgos Kordopatis{-}Zilos, Christos Tzelepis, Symeon Papadopoulos, Ioannis
  Kompatsiaris, and Ioannis Patras.
\newblock Dns: Distill-and-select for efficient and accurate video indexing and
  retrieval.
\newblock {\em International Journal of Computer Vision}, 130(10):2385--2407,
  2022.

\bibitem{w2vv++}
Xirong Li, Chaoxi Xu, Gang Yang, Zhineng Chen, and Jianfeng Dong.
\newblock {W2VV++}: Fully deep learning for ad-hoc video search.
\newblock In {\em ACMMM}, pages 1786--1794, 2019.

\bibitem{sea}
Xirong Li, Fangming Zhou, Chaoxi Xu, Jiaqi Ji, and Gang Yang.
\newblock {SEA}: Sentence encoder assembly for video retrieval by textual
  queries.
\newblock {\em TMM}, 23:4351--4362, 2021.

\bibitem{ce}
Yang Liu, Samuel Albanie, Arsha Nagrani, and Andrew Zisserman.
\newblock Use what you have: Video retrieval using representations from
  collaborative experts.
\newblock In {\em BMVC}, page 279, 2019.

\bibitem{ts2net}
Yuqi Liu, Pengfei Xiong, Luhui Xu, Shengming Cao, and Qin Jin.
\newblock {TS2-Net}: Token shift and selection transformer for text-video
  retrieval.
\newblock In {\em ECCV}, pages 319--335, 2022.

\bibitem{sgdr}
Ilya Loshchilov and Frank Hutter.
\newblock Sgdr: Stochastic gradient descent with warm restarts.
\newblock {\em arXiv}, 2016.

\bibitem{clip4clip}
Huaishao Luo, Lei Ji, Ming Zhong, Yang Chen, Wen Lei, Nan Duan, and Tianrui Li.
\newblock {CLIP4Clip}: An empirical study of {CLIP} for end to end video clip
  retrieval and captioning.
\newblock {\em Neurocomputing}, 508:293--304, 2022.

\bibitem{xclip}
Yiwei Ma, Guohai Xu, Xiaoshuai Sun, Ming Yan, Ji Zhang, and Rongrong Ji.
\newblock {X-CLIP}: End-to-end multi-grained contrastive learning for
  video-text retrieval.
\newblock In {\em ACMMM}, pages 638--647, 2022.

\bibitem{je}
Niluthpol~Chowdhury Mithun, Juncheng Li, Florian Metze, and Amit~K
  Roy-Chowdhury.
\newblock Learning joint embedding with multimodal cues for cross-modal
  video-text retrieval.
\newblock In {\em ICMR}, pages 19--27, 2018.

\bibitem{relation}
Wonpyo Park, Dongju Kim, Yan Lu, and Minsu Cho.
\newblock Relational knowledge distillation.
\newblock In {\em CVPR}, pages 3967--3976, 2019.

\bibitem{w2vvpp-bert}
Ladislav Pe\v{s}ka, Gregor Koval\v{c}\'{\i}k, Tom\'{a}\v{s} Sou\v{c}ek,
  V\'{\i}t \v{S}krh\'{a}k, and Jakub Loko\v{c}.
\newblock {W2VV++} {BERT} model at {VBS} 2021.
\newblock In {\em MMM}, 2021.

\bibitem{clip}
Alec Radford, Jong~Wook Kim, Chris Hallacy, Aditya Ramesh, Gabriel Goh,
  Sandhini Agarwal, Girish Sastry, Amanda Askell, Pamela Mishkin, Jack Clark,
  et~al.
\newblock Learning transferable visual models from natural language
  supervision.
\newblock In {\em ICML}, pages 8748--8763, 2021.

\bibitem{fitnets}
Adriana Romero, Nicolas Ballas, Samira~Ebrahimi Kahou, Antoine Chassang, Carlo
  Gatta, and Yoshua Bengio.
\newblock Fitnets: Hints for thin deep nets.
\newblock {\em arXiv}, 2014.

\bibitem{vatex}
Xin Wang, Jiawei Wu, Junkun Chen, Lei Li, Yuan-Fang Wang, and William~Yang
  Wang.
\newblock {VATEX}: A large-scale, high-quality multilingual dataset for
  video-and-language research.
\newblock In {\em ICCV}, pages 4581--4591, 2019.

\bibitem{msrvtt}
Jun Xu, Tao Mei, Ting Yao, and Yong Rui.
\newblock {MSR-VTT}: A large video description dataset for bridging video and
  language.
\newblock In {\em CVPR}, pages 5288--5296, 2016.

\bibitem{msrvtt-1ka}
Youngjae Yu, Jongseok Kim, and Gunhee Kim.
\newblock A joint sequence fusion model for video question answering and
  retrieval.
\newblock In {\em ECCV}, pages 471--487, 2018.

\bibitem{centerclip}
Shuai Zhao, Linchao Zhu, Xiaohan Wang, and Yi Yang.
\newblock {CenterCLIP}: Token clustering for efficient text-video retrieval.
\newblock In {\em SIGIR}, page 970–981, 2022.

\end{thebibliography}
}

\end{document}


\title{Supplementary Material for ``TeachCLIP: Multi-Grained Teaching for Efficient Text-to-Video Retrieval"}

\author{First Author\\
Institution1\\
Institution1 address\\
{\tt\small firstauthor@i1.org}
\and
Second Author\\
Institution2\\
First line of institution2 address\\
{\tt\small secondauthor@i2.org}
}

\maketitle
\ificcvfinal\thispagestyle{empty}\fi

In this supplementary material, we report extra results which are not included in the main paper. 

Table \ref{ab_afa} shows the influence of the number of fully connected (FC) layers in our Attentional frame-Feature Aggregation (AFA) module. Our current choice of using two FC layers is the best.



\begin{table}[ht!]
\caption{\textbf{Influence of the number of FC layers in AFA on the text-to-video retrieval performance}. Dataset: MSRVTT-1k. Teacher network: X-CLIP.} 
\label{ab_afa}
\setlength{\tabcolsep}{16pt} 
\renewcommand{\arraystretch}{1} 
\resizebox{\linewidth}{!}{
\begin{tabular}{@{}lrrrr@{}}
\toprule
\textbf{\#FC layers} & \textbf{R1} & \textbf{R5} & \textbf{R10} & \textbf{SumR} \\ \hline
1 & 44.9 & 72.2 & 81.9 & 199.0 \\
2 (used in the paper) & 45.2 & 72.3 & 82.3 & 199.8 \\
3 & 44.2 & 73.2 & 82.0 & 199.4 \\
\bottomrule
\end{tabular}
}
\end{table}


\stepcounter{magicrownumbers}
\begin{table*}[!h]
\centering
\caption{\textbf{Video-to-text retrieval performance on multiple datasets}. TeachCLIP uses X-CLIP as the teacher.
}
\label{table:v2t_alldatasets}
\renewcommand{\arraystretch}{1.1} 
\resizebox{\linewidth}{!}{
\begin{tabular}{@{}lrrrrrrrrrrrrrrr@{}}
\toprule
\multirow{2}{*}{\textbf{model}} & \multicolumn{3}{c}{\textbf{MSRVTT-1k}} & \multicolumn{3}{c}{\textbf{MSRVTT-3k}} & \multicolumn{3}{c}{\textbf{MSVD}} & \multicolumn{3}{c}{\textbf{VATEX}} & \multicolumn{3}{c}{\textbf{ActNetCap}} \\
\cmidrule(r){2-4} \cmidrule(r){5-7} \cmidrule(r){8-10} \cmidrule(r){11-13} \cmidrule(r){14-16}
 & R1 & R5 & SumR & R1 & R5 & SumR & R1 & R5 & SumR & R1 & R5 & SumR & R1 & R5 & SumR \\ \hline

CenterCLIP \cite{centerclip} & 42.8 & 71.7 & 196.7 & - & - & - & 57.9 & 83.6 & 232.0 & - & - & - & 44.5 & 75.7 & 206.4 \\
TS2-Net \cite{ts2net} & 43.6 & 71.1 & 197.4 & 53.0 & 82.5 & 225.3 & 62.5 & 85.8 & 240.9 & 78.7 & 97.8 & 275.8 & 41.3 & 72.1 & 198.5 \\
X-Pool \cite{xpool} & 45.1 & 73.7 & 202.6 & - & - & - & - & - & - & - & - & - & - & - & - \\ [3pt]

X-CLIP \cite{xclip} & 44.9 & 73.4 & 200.2 & 56.1 & 84.3 & 232.7 & 64.3 & 87.5 & 244.9 & 77.3 & 97.4 & 273.8 & 44.4 & 74.7 & 205.4 \\
CLIP4Clip \cite{clip4clip} & 41.4 & 70.6 & 192.5 & 50.8 & 78.5 & 216.3 & 63.3 & 85.4 & 240.4 &  77.0 & 97.4 & 273.4 & 39.5 & 71.4 & 194.5 \\ 
TeachCLIP & 43.6 & 71.3 & 196.5 & 54.8 & 82.8 & 228.5 & 63.5 & 86.4 & 241.8 & 80.0 & 98.4 & 277.9 & 41.9 & 73.9 & 201.5 \\
\bottomrule
\end{tabular}
}
\end{table*}

While aiming for text-to-video retrieval, we  also report video-to-text retrieval performance for a more complete picture of our evaluation, see Table \ref{table:v2t_alldatasets}. Results similar to the T2VR task can be observed. That is, TeachCLIP consistently reduces the performance gap between the student (CLIP4Clip) and the teacher (X-CLIP) on all datasets, while maintaining the same computation and storage efficiency as the student.


{\small
\bibliographystyle{ieee_fullname}
\bibliography{egbib}
}